\pdfoutput=1
\documentclass[11pt]{article}

\usepackage[]{ACL2023}

\usepackage{times}
\usepackage{latexsym}
\usepackage{amsmath}
\usepackage{amssymb}

\usepackage[labelfont=bf]{caption}
\usepackage[T1]{fontenc}
\usepackage[utf8]{inputenc}

\usepackage{microtype}

\usepackage{inconsolata}

\usepackage{graphicx}
\usepackage{subfig}
\usepackage{multirow}
\newcommand*\rot{\rotatebox{90}}
\usepackage{amssymb}
\usepackage{pifont}
\newcommand{\cmark}{\ding{51}}
\newcommand{\xmark}{\ding{55}}%

\usepackage{todonotes}

\title{A Match Made in Heaven: A Multi-task Framework for Hyperbole and Metaphor Detection}

\author{Naveen Badathala$^*$,
Abisek Rajakumar Kalarani$^*$, Tejpalsingh Siledar$^*$, \\  \bf{Pushpak Bhattacharya}\\
        Department of Computer Science and Engineering, IIT Bombay, India\\
        \texttt{\{naveenbadathala, abisekrk, tejpalsingh, pb\}@cse.iitb.ac.in}
        }

\begin{document}
\maketitle
\def\thefootnote{*}\footnotetext{Equal contribution.}
\def\thefootnote{1}\footnotetext{Code and data are available at: \url{https://github.com/abisekrk/multitask_hyperbole_metaphor_detection}}

% \thanks{}
\begin{abstract}
% Hyperbole and metaphor usage is common in everyday communication which makes their detection important. Previous approaches to hyperbole and metaphor detection have studied them independently but their relationship has never been explored. In this paper, we propose a multi-task deep learning framework to detect hyperbole and metaphor simultaneously. We hypothesize that metaphors help in hyperbole detection and vice versa. We manually annotate the Hypo dataset with labels for both hyperbole and metaphor. Experiments on the dataset show the efficacy of our proposed method for hyperbole detection gaining an improvement of 10\% over the current state-of-the-art systems. Our multi-task learning approach also shows an improvement of 4-5\% against the single-task learning approach for both hyperbole and metaphor detection. Our analysis further indicate the effectiveness of capturing the relationship between hyperbole and metaphors.

Hyperbole and metaphor are common in day-to-day communication ({\it e.g.}, "I am in deep trouble": how does {\it trouble} have {\it depth}?), which makes their detection important, especially in a conversational AI setting. Existing approaches to automatically detect metaphor and hyperbole have studied these language phenomena independently, but their relationship has hardly, if ever, been explored computationally. In this paper, we propose a multi-task deep learning framework to detect hyperbole and metaphor simultaneously. We hypothesize that metaphors help in hyperbole detection, and vice-versa. To test this hypothesis, we annotate two hyperbole datasets- {\it HYPO} and {\it HYPO-L}- with metaphor labels. Simultaneously, we annotate two metaphor datasets- {\it TroFi} and {\it LCC}- with hyperbole labels. Experiments using these datasets give an improvement of the state of the art of hyperbole detection by $\sim\mathbf{12\%}$. Additionally,  our multi-task learning (MTL) approach shows an improvement of up to $\sim\mathbf{17\%}$ over single-task learning (STL) for both hyperbole and metaphor detection, supporting our hypothesis. To the best of our knowledge, ours is the first demonstration of computational leveraging of linguistic intimacy between metaphor and hyperbole, leading to showing the superiority of MTL over STL  for hyperbole and metaphor detection$^1$.
\end{abstract}
% \footnote{\thanks{$*$Equal contributions from the authors}}
\section{Introduction}

% Plan:
% Start from explaining figurative language.
% Say why it is important
% Whine, hyperbole is not researched well
% Say, hyperbole and metaphor are related
% Highlight success of multi-task approaches in other figurative languages
% Mention contribution
% \todo{Increase the content - RK}

\begin{figure}
\centering
    \includegraphics[width=\linewidth]{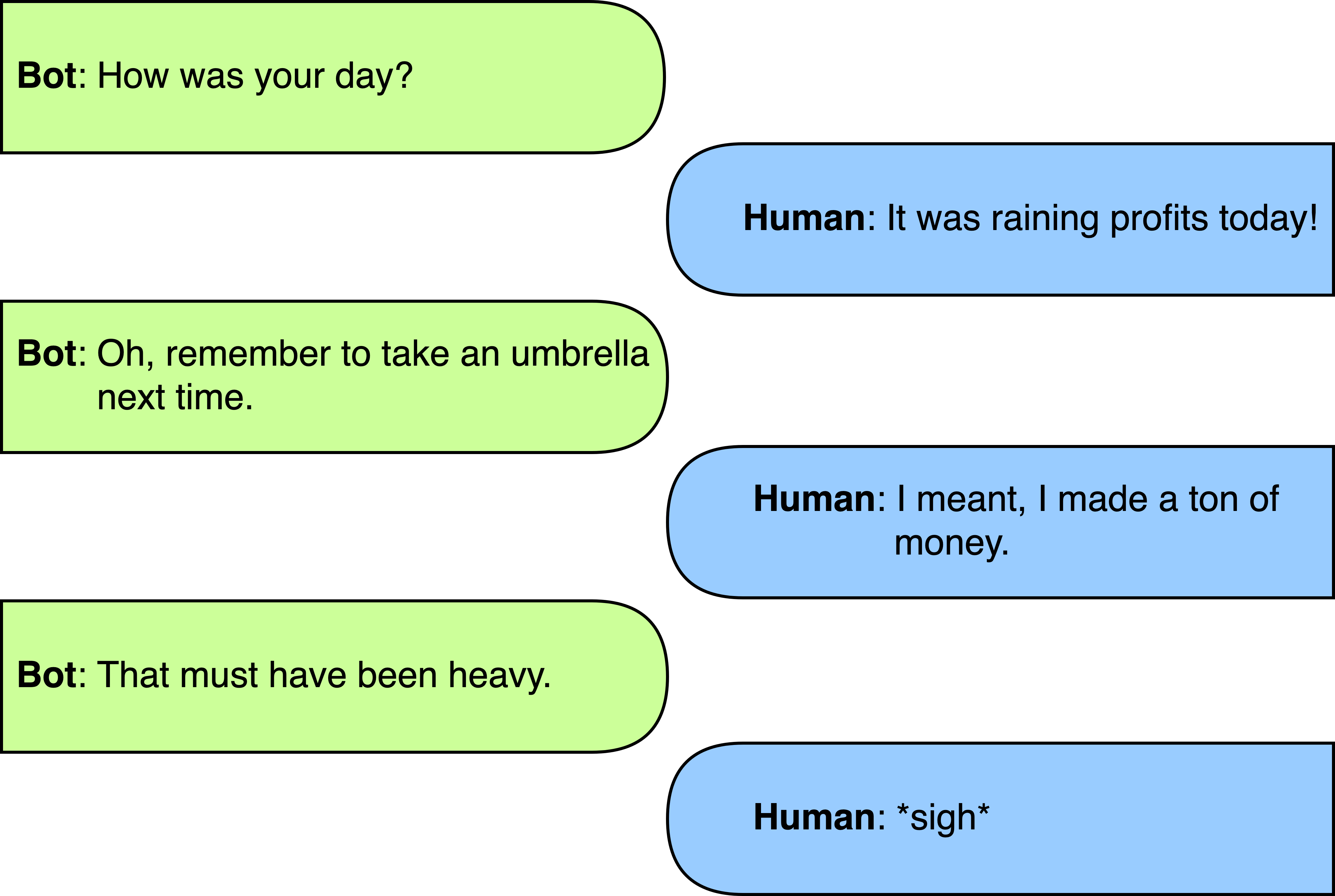}
    \caption{An example of the need for detecting hyperbolic and metaphoric sentences for AI systems.}
    \label{fig:chat example}
 \end{figure}

The use of figurative language is very common in natural discourse, and it is reflected in the content generated in social media networks \citep{10.1145/3375547}. Figurative languages are used to establish some communicative goals such as expressing a negative emotion, drawing attention to a part of the text, or adding interest to a subject. \citep{doi:10.1111/j.1467-9280.1994.tb00653.x}. The understanding of figurative languages like sarcasm, metaphor, simile, irony, and hyperbole is crucial for many NLP tasks such as building accurate sentiment analysis systems or developing conversational AI systems that can hold meaningful conversations (Figure \ref{fig:chat example}). This has led to great interest and value in understanding these figurative languages. Figurative languages like metaphor \citep{10.1145/3373265} and sarcasm \citep{joshi2017automatic} are studied extensively while hyperbole remains less explored. 

%\section{Motivation}

\begin{figure*}
    \centering
    \hspace*{0.5cm}
    \includegraphics[scale=0.5]{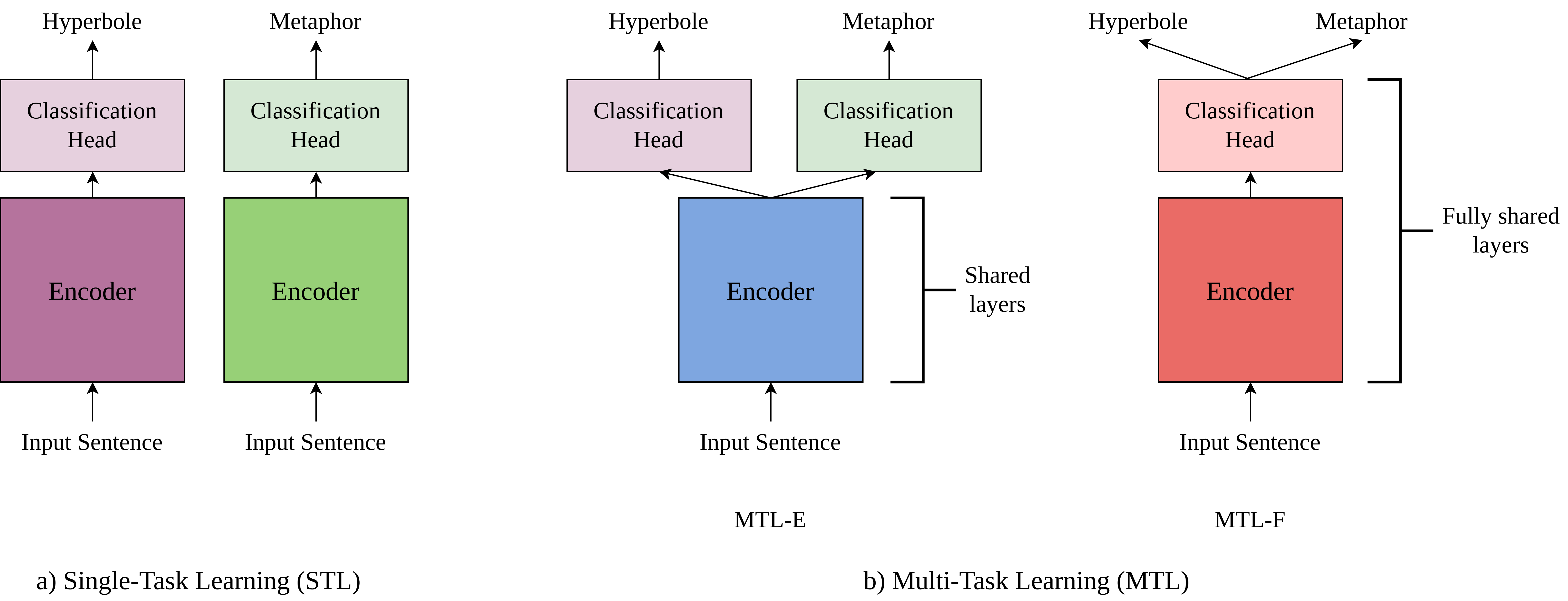}
    \caption{Overview of the single-task and multi-task learning architectures for detecting hyperbole and metaphor. a) Single-Task Learning (STL) model. b) Two variants of the Multi-Task learning (MTL) model: Multi-task learning with shared encoders (MTL-E) model and Multi-Task Learning with fully shared layers (MTL-F) model.}
    \label{fig:overview}
\end{figure*}

Metaphor is the most common choice of figurative language, while hyperbole is the second most adopted rhetorical device in communication \citep{1996figcooccurence} and hence it is important to study and process them automatically. Hyperbole is a figurative language that uses exaggeration to emphasize a point, while metaphor makes a comparison between two things to indicate a resemblance. 

% Metaphors use comparison of objects or ideas to indicate the likeliness between them. Hyperbole is an exaggerated version of a statement often used for emphasis.

\subsection{Motivation}
Relevance theorists had long treated both metaphors and hyperboles as not genuinely distinct categories as they are very closely related to each other \citep{Sperber200884}. Recent research has highlighted the distinctive features of hyperboles over metaphors \citep{CARSTON201579}. However, on the computational side, hyperbole and metaphor detection have been treated as isolated problems so far. 

Both metaphors and hyperboles use figurative elements to express an idea rather than presenting them literally, but this linguistic insight hasn't been exploited computationally in previous works.
% Metaphors use elements that express the same original idea in a different way using similar ideas whereas hyperboles use exaggeration to express the same but with more emphasis. 
% As both of these phenomena express the same idea using some other element,
We hypothesize that this shared characteristic can be captured at the embedding level by training transformer models to learn these representations jointly using multi-task learning. Existing metaphor detection systems focus on identifying metaphoricity at the token-level, whereas hyperbole detection systems focus on sentence-level classification. In our work, we highlight the effectiveness of performing sentence-level classification for both hyperboles and metaphors in a multi-task setting.

% Metaphors and hyperboles both use figurative elements to express an idea, rather than stating it literally. As they share this similarity, we believe that using multi-task learning to train transformer models on both could capture this commonality at the embedding level.

% \begin{figure*}
%     \centering
%     \hspace*{0.5cm}
%     \includegraphics[scale=0.5]{images/architecture_latest.png}
%     \caption{An overview of the single task and multi-task approaches for detecting hyperbole and metaphor.}
%     \label{fig:overview}
% \end{figure*}

\subsection{Contributions}
% In this work, we investigate the relationship between hyperbole and metaphor by formulating a novel multi-task detection system that detects the presence of both hyperbole and metaphor in a sentence. 
Our contributions are:
% \begin{itemize}
%     \item We perform sentence-level and token-level annotations and release extensions of the existing hyperbole datasets- HYPO and HYPO-L by adding metaphor labels to them and existing metaphor datasets- TroFi and LCC by adding hyperbole labels to them, accounting to $16,024$ sentences in total.
%     \item We propose and demonstrate the efficacy of our multi-task deep learning framework over a  single-task deep learning framework for hyperbole and metaphor detection.
%     \item We present the state-of-the-art for hyperbole detection on the HYPO dataset and metaphor detection on the LCC dataset using our multi-task learning framework.
%     % to detect hyperboles and metaphors simultaneously 
% \end{itemize}

\begin{enumerate}
    \item Extensions to the existing datasets amounting to $\textbf{16, 024}$ sentences which include,
    \begin{enumerate}
        \item HYPO and HYPO-L datasets annotated with metaphor labels.
        \item TroFi and LCC datasets annotated with hyperbole labels.
    \end{enumerate}

    \item Demonstration of the superiority of multitasking over single-tasking for hyperbole and metaphor detection.
    \item State-of-the-art results for sentence-level hyperbole detection on the HYPO dataset (F1 score- $\textbf{0.881}$).
    \item Benchmark results for sentence-level metaphor detection on our label-balanced LCC dataset (F1 score- $\textbf{0.805}$).
    % and metaphor detection on the LCC dataset using our multi-task learning framework.
    % to detect hyperboles and metaphors simultaneously 
\end{enumerate}

%We publish our datasets and code demos to foster further research \footnote{\url{https://github.com/abisekrk/multitask_hyperbole_metaphor_detection}}.

\begin{figure}
\centering
    \includegraphics[width=\linewidth]{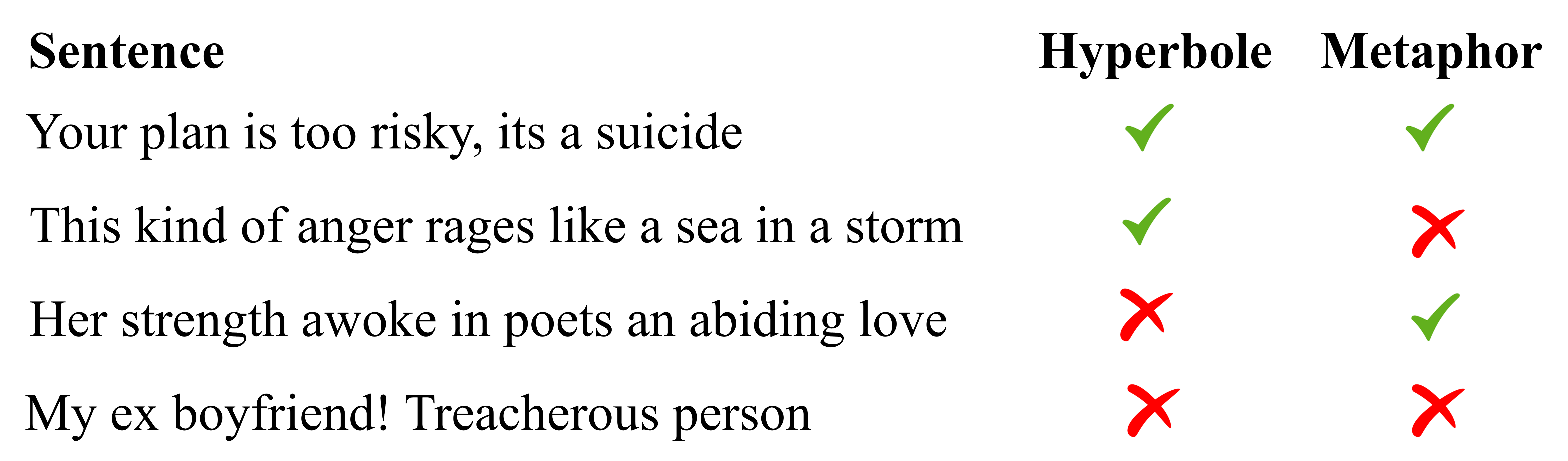}
    \caption{Example sentences with Hyperbole and Metaphor labels.}
    \label{fig:examples}
 \end{figure}

\section{Background and Definitions}

% \todo{Abisek: intro and examples of hyperbole and metaphor}
% \todo{Add table for examples}
% In this section, we formally define the figurative languages used in our work. 
\paragraph{Metaphor}
% A metaphor is a figure of speech that makes an implicit comparison to something that is literally not true. 
Metaphor is a literary device that uses an implicit comparison to drive home a new meaning.
Metaphors consist of a source and target domain in which the features from the source domain are related to the features in the target domain through comparable properties \citep{lakoff1993contemporary}. For instance, \emph{"Life is a journey,"} implies a comparison between life and journey through the idea of having a beginning and an end.
% The source domain 'journey' represents the property of having a start and an end and maps it to the target domain 'life' making an implicit comparison. 
In this work, we do not consider similes as metaphors as they make an explicit comparison.
% Here, the source domain `\emph{journey}' has a defined start and end as the property and it is mapped to the target domain `\emph{life}', bringing out an implicit comparison between life and journey through the property of having a start and an end.

% \paragraph{Simile}
% A simile is a figure of speech where two unrelated things are compared explicitly \citep{israel2004simile}. Similes are explicit about the comparison, whereas metaphors have a subtlety associated with them giving them more flexibility. For instance, in \emph{``He fought like a lion"}, a man is compared with a lion explicitly which could be expressed with a metaphor \emph{``He is a lion"}. However, the metaphor \emph{``His judgement is somewhat murky"} cannot be expressed accurately with a simile as \emph{``His judgement is something like murky"}.

% For Hyperbole

% add qualitative, quantitive directions
\paragraph{Hyperbole}
Hyperbole is a figurative language in which the literal meaning is exaggerated intentionally. It exaggerates expressions and blows them up beyond the point they are perceived naturally with the objective of emphasizing them \citep{claridge2010hyperbole}.
% Here, exaggeration can be brought about quantitatively by increasing or decreasing the quantity of the object or qualitatively by changing the subjective property of the object \citep{mora2009all}. 
% It often makes use of similes and metaphors for bringing out exaggeration but it is not mandatory. % Consider the following sentences:
% \begin{enumerate}
%     \item \textit{I'm tired, I can't lift my hand.}
%     \item \textit{My heart is bleeding right now.}
%     \item \textit{Her anger radiated like a nuclear explosion.}
% \end{enumerate}
% In the first sentence, the phrase, ``can't lift my hand" is an exaggeration. The exaggeration here is brought about without any comparison. 
For example, \textit{"I'm tired, I can't lift my hand,"} exaggerates the speaker's exhaustion. Figure \ref{fig:examples} shows examples of metaphor and hyperbole.
% \textit{"My heart is bleeding right now,"} exaggerates the speaker's emotions by using a metaphor and 
% the exaggeration is achieved with the help of a metaphor as we make an implicit comparison to bleeding, to drive home our point. 
% In the third example, anger is explicitly compared to a nuclear explosion to underline exaggeration. 
% \textit{"Her anger radiated like a nuclear explosion,"} exaggerates the speaker's anger by using an explicit comparison to a nuclear explosion. 
% In our work, 
% % we work with all these type of hyperboles and 
% we study the impact of understanding the metaphoricity of statements in identifying hyperboles of all these types.

% For example, consider the sentence "I will walk a thousand mile to meet you". Here   In the sentence "My heart is bleeding right now",  In the former example of walking thousand miles, the exaggeration was brought out without using any metaphors.

% \section{Motivation}

% \section{Task Formulation and Contributions}
% Our contributions include:

% \begin{figure}
%     \centering
%     \includegraphics[scale=0.45]{images/final_cr.png}
%     % \includegraphics[width=0.5\textwidth]{EMNLP 2022/images/emnlp_front_page_example_2.png}
%     \caption{An overview of our multi-task approach for detecting hyperbole and metaphor simultaneously.}
%     \label{fig: overview}
% \end{figure}

\section{Related Work}
% \subsection*{Hyperbole}
% \todo{Increase the content}

Metaphors and hyperboles are the most used figures of speech in everyday utterances \citep{1996figcooccurence}. In recent years, significant efforts have been made to understand metaphors and hyperboles, giving rise to interesting techniques to automatically detect and generate them. \citet{troiano2018computational} introduced hyperbole detection as a binary classification task, using traditional machine learning algorithms. They also released a dataset named `HYPO' for hyperbole detection. 
% They used a feature set composed of imageability, unexpectedness, polarity, subjectivity, and emotional intensity. The classification was done with traditional machine learning algorithms.
% like logistic regression, svm, and decision trees
% \todo{add references}. 
\citet{kong-etal-2020-identifying} introduced `HYPO-cn', a Chinese dataset for hyperbole detection, and showed that deep learning models can perform better at hyperbole detection with increased data. \citet{biddle-etal-2021-harnessing} used a BERT \citep{devlin2018bert} based detection system that used the literal sentences of the hyperbolic counterparts to identify the hyperbolic and non-hyperbolic use of words and phrases. They also released a test suite for evaluating models. \citet{tian-etal-2021-hypogen-hyperbole} proposed a hyperbole generation task.  \citet{zhang-wan-2022-mover} introduced an unsupervised approach for generating hyperbolic sentences from literal sentences and introduced two new datasets `HYPO-XL' and `HYPO-L' for their experiments.

Metaphors have been extensively studied even before hyperbole detection was introduced. \citet{tsvetkov-etal-2014-metaphor} introduced the TSV dataset with $884$ metaphorical and non-metaphorical adjective-noun (AN) phrases. They showed that conceptual mapping learnt between literal and metaphorical words is transferable across languages. \citet{mohler-etal-2016-introducing} introduced the LCC dataset which contains sentence-level annotations for metaphors in four languages totaling $188,741$ instances. \citet{Steen2010AMF} studied metaphor at the word level and was the first to include function words for metaphor detection with the new VUA dataset. \citet{birke-sarkar-2006-clustering} introduced the TroFi dataset that consists of verbs in their literal and metaphoric form. In recent years, metaphor detection has been explored with the aid of large language models. \citet{Choi2021MelBERTMD} used the contextual embeddings from BERT \cite{devlin2018bert} and RoBERTa \cite{Liu2019RoBERTaAR} to classify metaphorical sentences. \citet{met_recent} probed and analyzed the metaphorical knowledge gained by large language models by testing them on metaphor datasets across languages.

Previous research on metaphor and hyperbole detection typically treats these figurative language forms separately, despite their common properties.  
% To the best of our knowledge, our work is the first to combine both hyperbole and metaphor detection. 
In this work, we propose a multi-task approach that simultaneously detects both hyperboles and metaphors, and demonstrate that this approach outperforms individual detection tasks with experimental results and detailed analysis.

% . We also provide experimental evidence that our novel multi-task framework achieves state-of-the-art results on the hyperbole detection task.

% They proposed a machine-learning based classification approach using qualitative and quantitative features to identify whether the given sentence is hyperbole or not. \citet{kong2020identifying} proposed a neural-based classification approach along with the features proposed by \citet{troiano2018computational} and achieved better performance. \citet{tian2021hypogen} proposed a classifier by fine-tuning the pre-trained BERT \cite{devlin2018bert} models  to identify the hyperbolic sentences.

\section{Task Formulation}
% \todo{Legend}
% The task is to build a multi-task deep learning framework that takes a sentence as input and generates two binary labels for the sentence where one label denotes if the sentence is hyperbole and the other indicates if the sentence is a metaphor. For example, if the input sentence is \emph{``My heart is bleeding right now"}, the model should generate a label as ``1, 1'' indicating it is both hyperbole and metaphor. Figure \ref{fig:overview} depicts an overall view of the problem statement. 
For a sentence $\textbf{x}$ and a corresponding label $y$ or labels $y_{1},..,y_{k} \, (k>1)$, we can mathematically formulate the different learning tasks shown in Figure \ref{fig:overview} as:
\\
\noindent \textbf{Single Task Learning (STL)} 
\begin{align}
    y^* &= \underset{y \in \{0,1\}}{argmax} \, P(y|\textbf{x};\theta)\\
    P(y|\textbf{x};\theta) &= \rho(f(E(\textbf{x})))
\end{align}
where $E$ and $f$ represent the encoder and the feed-forward neural network (classification head) respectively, $\theta$ represents the weights from both $E$ and $f$, and $\rho$ represents the softmax function. The cross-entropy loss function can be defined as:
\begin{align}
    \mathcal{L} &= \frac{-1}{D} \sum_{i=1}^{D}(y_{i}\log\hat{y_{i}} + (1-y_{i})\log(1-\hat{y_{i}})) \label{eq1}
\end{align}
where $D$ is the number of training samples, $y_{i}$ and $\hat{y_{i}}$ are the $i^{th}$ true and predicted labels.\\
\noindent \textbf{Multi-Task Learning with shared Encoder (MTL-E)} 
\begin{align}
    y^*_{k} &= \underset{y_{k} \in \{0,1\}}{argmax} \,  P(y_{k}|\textbf{x};\theta_{k})\\
    % y^*_{2} &= \underset{y_{2} \in \{0,1\}}{argmax} P(y_{2}|\textbf{x};\theta_{2}) \\
    P(y_{k}|\textbf{x};\theta_{k}) &= \rho(f_{k}(E(\textbf{x})))
    % P(y_{2}|\textbf{x};\theta) &= softmax(f_{2}(E(\textbf{x})))
\end{align}
where $k$ represents the number of tasks, $y_{k}$ are the labels, $f_{k}$ are feed-forward neural networks and $\theta_{k}$ are the weights for the $k$ tasks respectively. For $k=2$ the loss function can be written as:
\begin{align}
    \mathcal{L} = \lambda\mathcal{L}_{1} + (1-\lambda)\mathcal{L}_{2}
\end{align}
where $\mathcal{L}_{1}, \mathcal{L}_{2}$ are task specific losses calculated similar to Eq. \ref{eq1} and $\lambda$ is the weighting factor.\\
\noindent \textbf{Multi-Task Learning with Fully shared layers (MTL-F)} 
\begin{align}
    y^*_{1},y^*_{2} &= \underset{y_{1},y_{2} \in \{0,1\}}{argmax} P(y_{1},y_{2}|\textbf{x};\theta)
    % P(y_{1},y_{2}|\textbf{x};\theta) &= \rho(f(E(\textbf{x})))
\end{align}
Here, the loss is a binary cross-entropy loss:
\begin{multline}
     \mathcal{L} = \frac{-1}{D} \sum_{i=1}^{D} \sum_{j=1}^{m}(y_{ij}\log(\sigma(l_{ij}) \\+ (1-y_{ij})\log(1-\sigma(l_{ij})))
\end{multline}
where $\sigma$ is the sigmoid function and $m$ is the number of labels and $l_{ij}$ represents the logit value for the $i^{th}$ instance and the $j^{th}$ label.

\section{Datasets and Annotations}
In this section, we delve into the hyperbole and metaphor datasets used and their annotation details.
% \todo{Dataset - legend}
% \todo{Calculate IAA}
% \todo{Reformat table}
% \todo{Add percentage stats in writing part}
%\todo{Naveen: add relevant current dataset details}
\subsection{Hyperbole Datasets}
Our experiments used two hyperbole datasets: HYPO \citep{troiano2018computational} and HYPO-L \citep{zhang-wan-2022-mover}. The HYPO dataset contains $709$ hyperbolic sentences each with a corresponding paraphrased literal sentence resulting in $1,418$ sentences.
% and a sentence containing the hyperbolic words/phrases in a non-hyperbolic context. 
% We used the hyperbolic and paraphrased sentences from the dataset, resulting in 1418 sentences. 
The HYPO-L dataset includes $1,007$ hyperbolic sentences and $2,219$ paraphrased sentences. For each sentence in the HYPO and HYPO-L datasets, we added metaphor labels.
% The HYPO dataset consists of 709 hyperbolic sentences. Each sentence is accompanied by a paraphrased literal sentence and a sentence containing the hyperbolic words/phrases in a non-hyperbolic context. We extracted the hyperbolic and paraphrased sentences from the dataset which amounts to 1418 sentences. The HYPO-L dataset consists of 1007 hyperbolic sentences and 2219 paraphrased sentences. For each sentence in HYPO and HYPO-L datasets, we add metaphor labels.
Table \ref{tab:ann_hyp} shows the statistics of the annotated hyperbole datasets.

\begin{table}[h]
\centering
\begin{tabular}{|c|c|c|r|} 
\hline
\textbf{Dataset (\# sentences)}                                                          & \textbf{Hyp.} & \textbf{Met.} & \textbf{\# sent.}  \\ 
\hline
\multirow{4}{*}{\begin{tabular}[c]{@{}l@{}}HYPO \\(1,418)\end{tabular}}    & \cmark                & \cmark                & 515                   \\ 
\cline{2-4}
                                                                          & \cmark                 & \xmark                & 194                   \\ 
\cline{2-4}
                                                                          & \xmark                 & \cmark                & 107                   \\ 
\cline{2-4}
                                                                          & \xmark                 & \xmark                & 602                   \\ 
\hline
\multirow{4}{*}{\begin{tabular}[c]{@{}l@{}}HYPO-L\\(3,326)\\\end{tabular}} & \cmark                 & \cmark                & 237                   \\ 
\cline{2-4}
                                                                          & \cmark                 & \xmark                & 770                   \\ 
\cline{2-4}
                                                                          & \xmark                 & \cmark                & 19                    \\ 
\cline{2-4}
                                                                          & \xmark                 & \xmark                & 2,200                  \\
\hline
\end{tabular}
\caption{Statistics of annotated hyperbole datasets with metaphor labels, where Hyp. means hyperbole, Met. means metaphor, and \#sent is the number of sentences.}
    \label{tab:ann_hyp}
\end{table}

\subsection{Metaphor Datasets}
We used two metaphor datasets: LCC \citep{mohler-etal-2016-introducing} and TroFi \citep{birke-sarkar-2006-clustering}. We manually annotated $3,838$ (out of $5,482$) sentences in the TroFi dataset and $7,542$ (out of $40,138$) sentences in the LCC dataset with hyperbole labels.
% We manually annotated 3838 sentences in TroFi dataset and 7542 sentences in LCC dataset with hyperbole labels. 
% Table \ref{tab:ann_met} shows the statistics of the annotated metaphor datasets. 
For statistics refer to Table \ref{tab:ann_met}.

\begin{table}
\centering
\begin{tabular}{|c|c|c|r|} 
\hline
\textbf{Dataset (\# sentences)}                                                       & \textbf{Met.} & \textbf{Hyp.} & \textbf{\# sent.}  \\ 
\hline
\multirow{4}{*}{\begin{tabular}[c]{@{}l@{}}TroFi\\(3,838)\end{tabular}} & \cmark           & \cmark           & 209                \\ 
\cline{2-4}
                                                                       & \cmark           & \xmark            & 1,710               \\ 
\cline{2-4}
                                                                       & \xmark            & \cmark           & 235                \\ 
\cline{2-4}
                                                                       & \xmark            & \xmark            & 1,684               \\ 
\hline
\multirow{4}{*}{\begin{tabular}[c]{@{}l@{}}LCC\\(7,542)\end{tabular}}   & \cmark           & \cmark           & 615                \\ 
\cline{2-4}
                                                                       & \cmark           & \xmark            & 3,187               \\ 
\cline{2-4}
                                                                       & \xmark            & \cmark           & 144                \\ 
\cline{2-4}
                                                                       & \xmark            & \xmark            & 3,596               \\
\hline
\end{tabular}
\caption{Statistics of annotated metaphor datasets with hyperbole labels, where Hyp. means hyperbole, Met. means metaphor, and \#sent is the number of sentences.}
    \label{tab:ann_met}
\end{table}

\subsection{Annotation Details} \label{subsection: annotation}
% We employed four annotators proficient in English. Three annotators were Masters students and one has an M.A in linguistics. All four annotators were male lying in the age group of 24-30. They were provided with detailed annotation instructions with examples of hyperbole and metaphors. To ensure the quality of annotations, we randomly sampled 500 instances with equal split of hyperbole and metaphor labels. 

We employed four 
%Master's students
annotators proficient in English in the age group of 24-30. Three annotators were master's students and one had an M.A in linguistics. They were provided with detailed annotation instructions along with examples of hyperbole and metaphors. Each instance of the dataset was annotated once and the annotations were equally divided among the four annotators. We first conducted pilot studies for annotation with randomly sampled 100 sentences from each dataset before proceeding to the final annotation. 
% More details in \textbf{Appendix} \ref{appendix: annotator details}. 
The Inter Annotator Agreement (IAA) was computed using pairwise Cohen's Kappa score ($\kappa$) and Fleiss' Kappa score (\textit{K}) as reported in Table \ref{tab:kappa}. The IAA between any two annotators is above $0.60$ ($ 0.61 \leq \kappa \leq 0.80;$ \citet{cohen1960coefficient}), indicating \textit{substantial} agreement between them. The Fleiss' Kappa score of $0.674$ is also considered \textit{substantial} 
($0.61\leq\textit{K}\leq 0.80;$ \citet{landis1977measurement}).

To ensure the quality of annotations, we randomly sampled $1100$ instances with an equal split of hyperbole and metaphor labels across all datasets. The annotators were asked to mark sentences as hyperbole if there was any exaggeration and as metaphors if there were any implicit comparisons. In addition to giving binary labels, we also asked the annotators to mark the part of the sentence that influenced their decisions. Doing this helped us identify any discrepancies in their understanding and correct them. All four annotators received stipends suitable for the tasks. 

% \begin{table}[h]
%     \centering
%     \begin{tabular}{|l|c|}
%     \hline
%     \multicolumn{2}{|c|}{\textbf{Pairwise Agreement}} &
%     \hline
%     Annotators A and B & 0.764\\
%     Annotators B and C & 0.661 \\
%     Annotators C and A & 0.837\\
%     Average pairwise Cohen's Kappa & 0.754\\
%     \hline
%     Fleiss' Kappa  & 0.753\\
%     \hline
%     % Fleiss' Kappa & --\\
%     % \hline
%     \end{tabular}
%     \caption{Inter-Annotator Agreement (Fleiss’ Kappa) values along with pairwise agreement among the annotators}
%     \label{tab:kappa}
% \end{table}

% \begin{table}
% \centering
% \begin{tabular}{|l|l|l|r|} 
% \hline
% \textbf{Cohen’s Kappa} & \textbf{A}                 & \textbf{B}                 & \multicolumn{1}{l|}{\textbf{C}}  \\ 
% \hline
% \textbf{B}             & \multicolumn{1}{r|}{0.749} & \multicolumn{1}{r|}{0.695} & 0.740                            \\ 
% \hline
% \textbf{B}             &                            & \multicolumn{1}{r|}{0.627} & 0.667                            \\ 
% \hline
% \textbf{C}             &                            &                            & 0.650                            \\ 
% \hline
% % \multicolumn{3}{|l|}{\textbf{Average pairwise Cohen's Kappa}}                    & \multicolumn{1}{l|}{0.688}       \\ 
% \hline
% \multicolumn{3}{|l|}{\textbf{Fliess’ Kappa}}                                     & 0.688                            \\
% \hline
% \end{tabular}
% \caption{Inter-Annotator Agreement calculations with Fleiss’ Kappa and pairwise Cohen’s Kappa among the annotators}
% \label{tab:kappa}
% \end{table}

\begin{table}[]
\centering
\begin{tabular}{|l|l|l|l|}
\hline
\textbf{Cohen's Kappa ($\kappa$)}   & A      & B      & C     \\ \hline
B               & 0.740  &        &       \\ \hline
C               & 0.651  & 0.653  &       \\ \hline
D               & 0.647  & 0.650  & 0.707 \\ \hline \hline
\multicolumn{3}{|l|}{\textbf{Fleiss' Kappa (\textit{K})}} & 0.674 \\\hline
\end{tabular}
\caption{IAA calculations with Fleiss’ Kappa and pairwise Cohen’s Kappa among the annotators}
\label{tab:kappa}
\end{table}

\begin{table*}[h]
    \centering
    \begin{tabular}{|c|l|c|c|c|c|c|c|}
    \hline
       \multirow{2}{*}{\textbf{Task}} & \multirow{2}{*}{\textbf{Model}} & \multicolumn{3}{c|}{\textbf{Hyperbole}} & \multicolumn{3}{c|}{\textbf{Metaphor}}\\
    \cline{3-8}
        & &  \small{Precision} & \small{Recall} & \small{F1} & \small{Precision} & \small{Recall} & \small{F1}\\
    \hline
    \hline
    \multirow{3}{*}{\small{STL}} 
    % BERT\textsubscript{base} & 0.788 & 0.735 & 0.756 & 0.720 & 0.631 & 0.670\\
     & \small{BERT\textsubscript{lg}} & 0.827 & 0.801 & 0.811 & 0.751 & 0.686 & 0.711\\
     & \small{ALBERT\textsubscript{xxl2}} & 0.845 & 0.871 & 0.856 & 0.695 & 0.736 & 0.713\\
     & \small{RoBERTa\textsubscript{lg}} & \textbf{0.883} & 0.848 & 0.864 & 0.801 & 0.709 & 0.745\\
    %  & deBERTa\textsubscript{large} & \textbf{0.899} & \textbf{0.887} & \textbf{0.892} & 0.797 & 0.741 & 0.766 \\
     \hline
     \multirow{3}{*}{\small{MTL-F}} 
    %  & BERT\textsubscript{base} & 0.818 & 0.785 & 0.792 & 0.790 & 0.645 & 0.706\\
      & \small{BERT\textsubscript{lg}} & 0.853 & 0.824 & 0.836 & 0.799 & 0.686 & 0.729\\
     & \small{ALBERT\textsubscript{xxl2}} & 0.847 & 0.878 & 0.860 & 0.757 & \textbf{0.761} & 0.753\\
     & \small{RoBERTa\textsubscript{lg}} & 0.879 & \textbf{0.884} & \textbf{0.881\textsuperscript{$\ast$}} & \textbf{0.826} & 0.752 & \textbf{0.787}\\
    %  & deBERTa\textsubscript{large} & 0.892 & 0.886 & 0.887 & \textbf{0.825} & 0.724 & \textbf{0.769} \\

     \hline
     
    \end{tabular}
    \caption{Comparison of Transformer models using 10-fold cross-validation over three different runs for hyperbole and metaphor detection task on the \textbf{HYPO} dataset. Significance test (t-test) p-value ($\ast$) = $0.0322$ (<$ 0.05$).}
    \label{tab:hypo_result}
\end{table*}

\begin{table*}[h]
    \centering
    \begin{tabular}{|c|l|c|c|c|c|c|c|}
    \hline
       \multirow{2}{*}{\textbf{Task}} & \multirow{2}{*}{\textbf{Model}} & \multicolumn{3}{c|}{\textbf{Hyperbole}} & \multicolumn{3}{c|}{\textbf{Metaphor}}\\
    \cline{3-8}
        & &  \small{Precision} & \small{Recall} & \small{F1} & \small{Precision} & \small{Recall} & \small{F1}\\
    \hline
    \hline
    \multirow{3}{*}{\small{STL}} & \small{BERT\textsubscript{lg}} & 0.670 & 0.598 & 0.630 & 0.561 & 0.466 & 0.506 \\
    & \small{ALBERT\textsubscript{xxl2}} & 0.649 & 0.542 & 0.589 & 0.513 & 0.414 & 0.456  \\
    & \small{RoBERTa\textsubscript{lg}} & 0.688 & 0.651 & 0.667 & 0.591 & 0.543 & 0.563\\
    % & deBERTa\textsubscript{large} & 0.697 & 0.649 & 0.670 (0.03) & - & - & -\\
    \hline
    \multirow{3}{*}{\small{MTL-F}} &
    \small{BERT\textsubscript{lg}} & 0.655 & 0.619 & 0.638 & 0.552 & 0.464 & 0.503 \\
    & \small{ALBERT\textsubscript{xxl2}} & 0.638 & 0.593 & 0.614 & 0.498 & 0.385 & 0.430 \\
    & \small{RoBERTa\textsubscript{lg}} & \textbf{0.706} & \textbf{0.668} & \textbf{0.687\textsuperscript{$\ast$}} & \textbf{0.599} & \textbf{0.554} & \textbf{0.572}\\
    % & deBERTa\textsubscript{large} & 0.709 & 0.664 & 0.685 (0.04) & 0.588 & 0.545 & 0.563 (0.04)\\
     \hline
     
    \end{tabular}
    \caption{Comparison of Transformer models using 10-fold cross-validation over three different runs for hyperbole and metaphor detection task on the \textbf{HYPO-L} dataset. Significance test (t-test) p-value ($\ast$) = $0.0438$ (< $0.05$).}
    \label{tab:hypol}
\end{table*}

\begin{table*}[h]
    \centering
    \begin{tabular}{|c|l|c|c|c|c|c|c|}
    \hline
       \multirow{2}{*}{\textbf{Task}} & \multirow{2}{*}{\textbf{Model}} & \multicolumn{3}{c|}{\textbf{Hyperbole}} & \multicolumn{3}{c|}{\textbf{Metaphor}}\\
    \cline{3-8}
        & &  \small{Precision} & \small{Recall} & \small{F1} & \small{Precision} & \small{Recall} & \small{F1}\\
    \hline
    \hline
    \multirow{3}{*}{\small{STL}} & \small{BERT\textsubscript{lg}} & 0.557 & 0.412 & 0.466 & 0.531 & 0.559 & 0.538\\
    & \small{ALBERT\textsubscript{xxl2}} & 0.424 & 0.234 & 0.294 & 0.489 & 0.430 & 0.454\\
    & \small{RoBERTa\textsubscript{lg}} & \textbf{0.607} & 0.446 & 0.496 & 0.542 & 0.469 & 0.490\\
    \hline
    \multirow{3}{*}{\small{MTL-F}} & \small{BERT\textsubscript{lg}} & 0.565  & 0.433 & 0.486 & 0.556 & 0.525 & 0.540\\
    & \small{ALBERT\textsubscript{xxl2}} & 0.487 & 0.241 & 0.312 & 0.516 & 0.457 & 0.475\\
    & \small{RoBERTa\textsubscript{lg}} & 0.605 & \textbf{0.529} & \textbf{0.561} & \textbf{0.565} & \textbf{0.587} & \textbf{0.573\textsuperscript{$\ast$}}\\
     \hline
     
    \end{tabular}
    \caption{Comparison of Transformer models using 10-fold cross-validation accuracy over three different runs for hyperbole and metaphor detection on the label balanced \textbf{TroFi} dataset. Significance test (t-test) p-value ($\ast$) < $0.0001$.}
    \label{tab:trofi_result}
\end{table*}

\begin{table*}[h]
    \centering
    \begin{tabular}{|c|l|c|c|c|c|c|c|}
    \hline
       \multirow{2}{*}{\textbf{Task}} & \multirow{2}{*}{\textbf{Model}} & \multicolumn{3}{c|}{\textbf{Hyperbole}} & \multicolumn{3}{c|}{\textbf{Metaphor}}\\
    \cline{3-8}
        & &  \small{Precision} & \small{Recall} & \small{F1} & \small{Precision} & \small{Recall} & \small{F1}\\
    \hline
    \hline
    \multirow{3}{*}{\small{STL}} & \small{BERT\textsubscript{lg}} & 0.649 & 0.542 & 0.589 & 0.758 & 0.736 & 0.745\\
    & \small{ALBERT\textsubscript{xxl2}} & 0.591 & 0.546 & 0.564 & 0.723 & 0.757 & 0.739\\
    & \small{RoBERTa\textsubscript{lg}} & \textbf{0.692} & 0.604 & 0.645 & \textbf{0.802} & 0.787 & 0.794\\
    \hline
    \multirow{3}{*}{\small{MTL-F}} & \small{BERT\textsubscript{lg}} & 0.633 & 0.531 & 0.575 & 0.750 & 0.774 & 0.760\\
    & \small{ALBERT\textsubscript{xxl2}} & 0.614 & 0.425 & 0.499 & 0.709 & 0.785 & 0.744\\
    & \small{RoBERTa\textsubscript{lg}} & 0.630 & \textbf{0.691} & \textbf{0.659} & 0.798 & \textbf{0.812} & \textbf{0.805\textsuperscript{$\ast$}}\\
     \hline
     
    \end{tabular}
    \caption{Comparison of Transformer models using 10-fold cross-validation over three different runs for hyperbole and metaphor detection on the label balanced \textbf{LCC}. Significance test (t-test) p-value ($\ast$) = $0.0221$ (< $0.05$).}
    \label{tab:lcc_result}
\end{table*}

\begin{table*}[h]
    \centering
    \begin{tabular}{|c|l|c|c|c|c|c|c|}
    \hline
       \multirow{2}{*}{\textbf{Task}} & \multirow{2}{*}{\textbf{Model}} & \multicolumn{3}{c|}{\textbf{Hyperbole}} & \multicolumn{3}{c|}{\textbf{Metaphor}}\\
    \cline{3-8}
        & &  \small{Precision} & \small{Recall} & \small{F1} & \small{Precision} & \small{Recall} & \small{F1}\\
    \hline
    \hline
    \multirow{1}{*}{\small{STL}} & 
    % \small{BERT\textsubscript{large}} & 0.827 & 0.801 & 0.811 & 0.758 & 0.736 & 0.745\\
    % & \small{ALBERT\textsubscript{large}} & 0.845 & 0.871 & 0.856 & 0.723 & 0.757 & 0.739\\
    % &
    \small{RoBERTa\textsubscript{lg}} & \textbf{0.883} & 0.848 & 0.864 & \textbf{0.802} & 0.787 & 0.794\\
    \hline
     \multirow{1}{*}{\small{MTL-E}} 
    %  & BERT\textsubscript{base} & 0.818 & 0.785 & 0.792 & 0.790 & 0.645 & 0.706\\
     %  & 
     %  \small{BERT\textsubscript{large}} & 0.817 & 0.788 & 0.802 & 0.737 & 0.762 & 0.749\\
     % & \small{ALBERT\textsubscript{large}} & 0.726 & 0.746 & 0.734 & 0.654 & 0.610 & 0.629\\
     & \small{RoBERTa\textsubscript{lg}} & 0.859 & 0.878 & 0.867 &  0.792 & 0.808 & 0.799\\
     \hline
    \multirow{1}{*}{\small{MTL-F}} 
    % & \small{BERT\textsubscript{large}} & 0.853 & 0.824 & 0.836 & 0.750 & 0.774 & 0.760\\
    % & \small{ALBERT\textsubscript{large}} & 0.847 & 0.878 & 0.860 & 0.709 & 0.785 & 0.744\\
    & \small{RoBERTa\textsubscript{lg}} & 0.879 & \textbf{0.884} & \textbf{0.881} & 0.798 & \textbf{0.812} & \textbf{0.805}\\
    \hline
     
    \end{tabular}
    \caption{Comparison of STL, MTL-E and MTL-F models using 10-fold cross-validation over three different runs on the \textbf{HYPO} dataset for hyperbole detection and the label balanced \textbf{LCC} dataset for metaphor detection. The metaphor column gives the \textbf{benchmark} results (sentence-level) on the label-balanced LCC dataset.}
    \label{tab:final_result}
\end{table*}

% \subsection{Annotation Guidelines}
% Before starting the annotation process the annotators were provided with proper guidelines and some examples. The annotators were asked to annotate each sentence of the HYPO dataset with a metaphor label. The definition mentioned in section 2 was used to determine whether a sentence is a metaphor or not.

\section{Experiments} \label{section: experiments}

% We perform four types of experiments: 1) A comparison between STL and MTL-F models on various hyperbole and metaphor datasets, 2) A comparison between STL, MTL-E and MTL-F models, 3) A comparison with established baseline models for the hyperbole dataset. 4) Benchmark results (sentence-level) on metaphor dataset.

We conduct four experiments: 1) Comparing STL and MTL-F on hyperbole and metaphor datasets, 2) Comparing STL, MTL-E, and MTL-F models, 3) Obtaining sentence-level benchmark results on the metaphor dataset, and 4) Comparing with established baselines for the hyperbole dataset.

For our experiments, we used label-balanced metaphor datasets to address the imbalance caused by fewer hyperbole (Refer to \textbf{Appendix} \ref{label-balanced}). To ensure a fair comparison, we used mean 10-fold cross-validation obtained over three different runs to compare our models. However, we did not compare our results with existing work on metaphor detection as it does token-level instead of sentence-level metaphor prediction. Finally, we used simple models to highlight the efficacy of a multi-tasked framework for a sophisticated task.

% We conduct two sets of experiments: 1) A comparison between single-task learning (STL) and multi-task learning (MTL) models over different hyperbole and metaphor datasets. 2) A comparison with existing baseline models for the hyperbole dataset.

% First, we formulated the task of hyperbole and metaphor detection as a supervised learning problem. Specifically, we carried out three tasks: a) Hyperbole detection (STL), which involves sentence-based classification with two classes: hyperbole and literal. b) Metaphor detection (STL), which involves sentence-based classification with two classes: metaphor and literal. c) Hyperbole and metaphor detection (MTL), which involves multi-label classification with two classes: metaphor and hyperbole. For our experiments, we used label-balanced metaphor datasets to address the imbalance caused by fewer hyperboles.

% Second, we compared the results of our models with existing baseline models using a mean 10-fold cross-validation accuracy obtained over three different runs for the hyperbole detection task using the \textbf{HYPO} dataset. We did not compare our results on metaphor datasets as existing work on metaphor detection does token-level instead of sentence-level metaphor prediction.  

\subsection{Hyperbole Baselines}
\citet{troiano2018computational} used cognitive features, such as imageability, unexpectedness, polarity, subjectivity, and emotional intensity for hyperbole detection,
% Imageability measures the degree to which a word evokes a mental image, as rated in the MRC psycholinguistic database. Unexpectedness is calculated as the average and maximum word pair similarity of a given sentence using word2vec and GloVe embeddings. Polarity expresses the sentiment of the sentence, subjectivity determines whether the sentence has an objective or personal opinion, and emotional intensity measures the strength of the sentiment. 
referred to as QQ (i.e. Qualitative and Quantitative). We compare our results with their best-performing Logistic Regression and Naive Bayes models, referred to as \textbf{LR+QQ} and \textbf{NB+QQ} in Table \ref{Table:baseline comparison}.

\citet{kong-etal-2020-identifying} used a combination of the QQ features and a pre-trained BERT, referred to as  \textbf{BERT\textsubscript{base}+QQ} in Table \ref{Table:baseline comparison}. The QQ features were concatenated with the BERT's output and passed through a linear classifier to predict hyperbole.

\citet{biddle-etal-2021-harnessing} used literal paraphrases as privileged information and incorporated this information using a triplet loss. We refer to this model as \textbf{BERT\textsubscript{base}+PI} in Table \ref{Table:baseline comparison}. We show that our multitask model outperforms all these baselines.
% \todo{hypo paper results}

% \todo{metaphor baseline}

\subsection{Experimental Setup} \label{setup}

We experiment with \verb|bert-large-uncased| (BERT\textsubscript{lg}) \cite{devlin2018bert}, \verb|albert-xxlarge-v2| (ALBERT\textsubscript{xxl2}) \cite{Lan2020ALBERTAL}, and  \verb|roberta-large| (RoBERTa\textsubscript{lg}) \cite{Liu2019RoBERTaAR} models ($h=16, l=24$).
%We use the weights and setting from the HuggingFace library for our models. 
The best-performing models use the following hyperparameters: For the STL model we use a learning rate of $1e-4$ for $5$ epochs and a batch size of $16$. For the MTL-E model, the learning rate is $1e-5$ for $20$ epochs, a batch size of $32$, and the loss weighting factor $\lambda$ of $0.5$ whereas, for the MTL-F model, the learning rate is $1e-5$ for $10$ epochs and a batch size of $16$. We use Adam \cite{Kingma2015AdamAM} with eps of $1e-4$ to optimize all our models. 
% Refer to \textbf{Appendix} \ref{hyperparameter}.
% The models trained to perform hyperbole detection were used to generate hyperbole labels for the metaphor datasets. We then perform multi-label classification for both hyperbole and metaphors on these augmented datasets and compare our results.
%  We did the hyperparameter search manually using: number of epochs = [5, 7, 10, 15, 20, 25],
% learning rate = [1e-5, 5e-5, 1e-4, 2e-4, 5e-4], and batch size = [4, 8, 16, 32, 64].
% % The following hyperparameter values were chosen based on the best-performing model:
% % Number of epochs: 5, Learning rate: 1e-4, Batch size: 16. We used 10-fold cross-validation for training and evaluation. 3 runs of such 10-fold cross-validations were performed. The training was performed for 5 epochs in each iteration.
% % Total number of training runs (overall): 10*5*3 = 150.
% The hyperparameters of the best-performing models have been mentioned in Section \ref{setup}. The training runs for STL, MTL-E, and MTL-F models were 150, 600, and 300 respectively and 30 evaluation runs each.

\subsection{Hyperparameter Details} \label{hyperparameter}
 % We use BERT, ALBERT, and RoBERTa based models for our experiments. For BERT and ALBERT models, the embedding for the CLS token is passed through a feed-forward neural network (FFN) for classification. For the RoBERTa-based model, the embedding from the <s> token is used in a similar fashion. For these models, embeddings for these special tokens represent the whole sentence rather than a pooling technique that generates the embedding for the sentence.

 We did hyperparameter search manually with the following search space: number of epochs = [5, 7, 10, 15, 20, 25],
learning rate = [1e-5, 5e-5, 1e-4, 2e-4, 5e-4], and batch size = [4, 8, 16, 32, 64].

The hyperparameters of the best-performing models have been mentioned in Section \ref{setup}. The training runs for STL, MTL-E, and MTL-F models were 150, 600, and 300 respectively and 30 evaluation runs each. 
% Compute details in Appendix \ref{section:appendix experiment}.

\subsection{Hypothesis Testing}\label{hypothesis_testing}
We used t-test, which is a statistical test used to determine if there is a significant difference between the means of two groups. The p-value here is a statistical measure that is used to assess the evidence against a null hypothesis. A p-value < 0.05 is typically considered to be statistically significant. The null hypothesis to reject here is that both the samples for STL and MTL-F models come from the same distribution. 

For all our experiments, we obtain a p-value < 0.05 indicating that the samples are indeed coming from different distributions. This shows that the improvement obtained by the MTL-F model over the STL model is statistically significant.

\begin{table}[h]
    \centering
    \begin{tabular}{|c|l|c|c|c|}
    \hline
        & \textbf{Model} & \textbf{P} & \textbf{R} & \textbf{F1}  \\
        \hline
        \hline
        \multirow{5}{*}{\textbf{\rot{Baselines}}} & \small{LR+QQ} 
        % \citet{troiano2018computational}  
        & 0.679 & 0.745 & 0.710\\
        & \small{NB+QQ} 
        % \citet{troiano2018computational} 
        & 0.689 & 0.696 & 0.693\\
        & \small{BERT\textsubscript{base}} 
        % \citet{kong-etal-2020-identifying}
        & 0.711 & 0.735 & 0.709\\
        & \small{BERT\textsubscript{base}+QQ}
        % \citet{biddle-etal-2021-harnessing} 
        & 0.650 & 0.765 & 0.671\\
        & \small{BERT\textsubscript{base}+PI} 
        % \citet{biddle-etal-2021-harnessing} 
        & 0.754 & 0.814 & 0.781\\
        \hline
        \hline
        \multirow{3}{*}{\textbf{\rot{Ours}}} 
        % & \small{BERT\textsubscript{large}-STL} & 0.827 & 0.801 & 0.811\\
        % & \small{ALBERT\textsubscript{large}-STL} & 0.845 & 0.871 & 0.856\\
        & \small{RoBERTa\textsubscript{lg} STL} & \textbf{0.883} & 0.848 & 0.864\\
        % & \small{BERT\textsubscript{large}-MTL} & 0.853 & 0.824 & 0.836\\
        % & \small{ALBERT\textsubscript{large}-MTL} & 0.847 & 0.878 & 0.860\\
        & \small{RoBERTa\textsubscript{lg} MTL-E} & 0.859 & 0.878 & 0.867\\
        & \small{RoBERTa\textsubscript{lg} MTL-F} & 0.879 & \textbf{0.884} & \textbf{0.881}\\
        \hline
        
    \end{tabular}
    \caption{\textbf{HYPO Results.} Precision (P), recall (R) and F1 score for baseline models compared to our work.}
    \label{Table:baseline comparison}
\end{table}

\section{Results}\label{section:results}

\begin{table*}[h]
    \centering
    \begin{tabular}{|l|c|c|c|c|}
        \hline
         \multirow{2}{*}{\textbf{ Sentences}} & \multirow{2}{*}{\textbf{Actual}} &
        \multirow{2}{*}{\textbf{MTL-F}} & \multicolumn{2}{c|}{\textbf{STL}}\\
        \cline{4-5}
        &  &  & \textbf{HD} & \textbf{MD}\\
        \hline
        \textit{Your plan is too risky, it's a suicide.} & H, M & H, M & \textcolor{red}{NH} & \textcolor{red}{NM}\\
        \hline
        \textit{I'm not staying here any longer!} & NH, NM & NH, NM & \textcolor{red}{H} & NM\\
        \hline
        \textit{This kind of anger rages like a sea in a storm.} & H, NM & H, NM & H & \textcolor{red}{M}\\
        \hline
        \textit{My ex boyfriend! Treacherous person!} & NH, NM & NH, NM & \textcolor{red}{H} & \textcolor{red}{M}\\
        \hline
        \textit{They cooked a turkey the size of a cow.} & H, M & H, M & H & \textcolor{red}{NM}\\
        \hline
        \textit{Her strength awoke in poets an abiding love.} & NH, M & NH, M & \textcolor{red}{H} & M\\
        \hline
        \textit{My sister is a vortex of intelligence in space.} & H, M & H, M & H & M\\
        \hline
        \textit{The act of love strongly resembles severe pain.} & NH, NM & NH, NM & NH & NM\\
        \hline
        
    \end{tabular}
    \caption{Some cases where the MTL-F performs better than the STL for hyperbole detection (HD) and metaphor detection (MD). Here H denotes a hyperbolic sentence, M denotes a metaphoric sentence, NH denotes a non-hyperbolic sentence, and NM denotes a non-metaphoric sentence. Notations in red indicate incorrect detection.}
    \label{tab:comparison_results}
\end{table*}

\begin{figure*}
    \centering
    \includegraphics[scale=0.4]{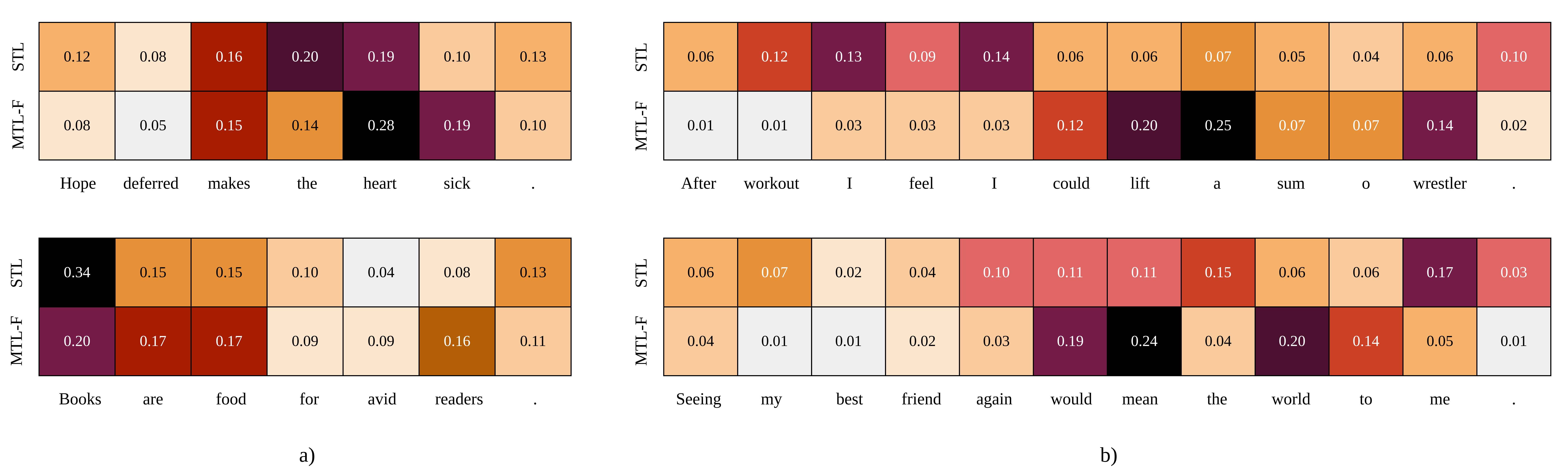}
    \caption{
    % Comparison of attention weights for both the STL and MTL models obtained by averaging over all the attention heads of the last layer for the classification token.
    Examples show the improvement in the focus of the MTL-F model over the STL model for two cases: a) Classifying a hyperbolic sentence in the presence of metaphor labels. b) Classifying a metaphoric sentence in the presence of hyperbole labels. (Darker colors indicate higher attention)}
    \label{fig:hyp_attention}
\end{figure*}

\begin{table*}[h]
    \centering
    \hspace*{-0.1cm}%
    \begin{tabular}{|l|c|c|}
        \hline
        \multirow{1}{*}{\textbf{ Sentences}} & \multirow{1}{*}{\textbf{Actual}} &
        \multirow{1}{*}{\textbf{MTL-F}}\\
        \hline
        \textit{What kind of sorcery is this?} & H, M & \textcolor{red}{NH}, \textcolor{red}{NM}\\
        \hline
        \textit{You're grumpy.} & NH, NM & NH, \textcolor{red}{M}\\
        \hline
        \textit{..this car is more a sophisticated piece of machinery than a regular car.} & NH, NM & \textcolor{red}{H}, NM\\
        \hline
        \textit{Stop bothering him: you're inviting trouble.} & NH, M & NH, \textcolor{red}{NM}\\
        \hline
        \textit{The work of the farm seemed to rest entirely on this horse's mighty shoulders.} & H, M & \textcolor{red}{NH}, M\\
        
        % \hline
        % 6 & \textit{TV is providing us with subliminal messages.} & H, M & \textcolor{red}{NH}, \textcolor{red}{NM}\\
        \hline
    \end{tabular}
    \caption{Error cases where MTL-F fails in the detection task. Here H denotes a hyperbolic sentence, M denotes a metaphoric sentence, NH denotes a non-hyperbolic sentence, and NM denotes a non-metaphoric sentence.}
    \label{tab:mtl_error}
\end{table*}

\textbf{STL vs. MTL-F models} \hspace{1em}We use identical experimental setups to compare the results obtained from the STL and MTL-F approach on all four datasets.

\textbf{1. HYPO results}:
The comparative analysis results for the HYPO dataset are in Table \ref{tab:hypo_result}. For all the models we observe that the MTL-F performs better than the corresponding STL. 
% The MTL BERT\textsubscript{base} model achieves an F1 of 0.792 (4.76\% $\uparrow$) for hyperbole and 0.706 (5.37\% $\uparrow$) for metaphor over the STL BERT\textsubscript{base} model with F1 of 0.756 for hyperbole and 0.670 for metaphor. The MTL BERT\textsubscript{large} model achieves an F1 of 0.836 (2.99\% $\uparrow$) for hyperbole and 0.729 (2.46\% $\uparrow$) for metaphor over the STL BERT\textsubscript{large} model with F1 of 0.811 for hyperbole and 0.711 for metaphor. The MTL ALBERT\textsubscript{large} model achieves an F1 of 0.860 (0.46\% $\uparrow$) for hyperbole and 0.753 (5.61\% $\uparrow$) for metaphor over the STL ALBERT\textsubscript{large} model with F1 of 0.856 for hyperbole and 0.713 for metaphor. The MTL RoBERTa\textsubscript{large} model achieves an F1 of 0.881 (1.96\% $\uparrow$) for hyperbole and 0.787 (5.33\% $\uparrow$) for metaphor over the STL RoBERTa\textsubscript{large} model with F1 of 0.864 for hyperbole and 0.745 for metaphor. 
Overall the RoBERTa\textsubscript{lg} MTL-F model achieves the best recall of $0.884$ and F1 of $\textbf{0.881}$ ($1.96$\% $\uparrow$) for hyperbole detection and a p-value of $\textbf{0.0322}$.

% as well as best precision of \textbf{0.826} and F1 of \textbf{0.787} for metaphor. result is statistically significant over the STL model with

% The ALBERT\textsubscript{large} model achieves the overall best recall of \textbf{0.761} for metaphor.

\textbf{2. HYPO-L results}:
The comparative analysis results for the HYPO-L dataset are in Table \ref{tab:hypol}. For all the models we observe that the MTL-F performs better than the corresponding STL for hyperbole detection. 
% The MTL BERT\textsubscript{large} model achieves an F1 of 0.638 (1.26\% $\uparrow$) for hyperbole and 0.503 (\% $\uparrow$) for metaphor over the STL BERT\textsubscript{large} model with F1 of 0.630 for hyperbole and - for metaphor. The MTL ALBERT\textsubscript{large} model achieves an F1 of - (\% $\uparrow$) for hyperbole and - (\% $\uparrow$) for metaphor over the STL ALBERT\textsubscript{large} model with F1 of - for hyperbole and - for metaphor. The MTL RoBERTa\textsubscript{large} model achieves an F1 of 0.687  (2.99\% $\uparrow$)for hyperbole and 0.572 (1.59\% $\uparrow$) for metaphor over the STL RoBERTa\textsubscript{large} model with F1 of 0.667 for hyperbole and 0.563 for metaphor. 
Overall the RoBERTa\textsubscript{lg} MTL-F model achieves the best precision of $0.706$, recall of $0.668$, and F1 of $\textbf{0.687}$  ($2.99$\% $\uparrow$) for hyperbole detection and a p-value of $\textbf{0.0438}$.
% as well as best precision of \textbf{0.599}, recall of \textbf{0.554}, and F1 of \textbf{0.572} for metaphor.

\textbf{3. TroFi results:}
The comparative analysis results for the label-balanced TroFi dataset is in Table \ref{tab:trofi_result}. For all the models we observe that the MTL-F performs better than the corresponding STL for metaphor detection. 
% The MTL BERT\textsubscript{large} model achieves an F1 of 0.486 (4.29\% $\uparrow$) for hyperbole and 0.540 (0.37\% $\uparrow$) for metaphor over the STL BERT\textsubscript{large} model with F1 of 0.466 for hyperbole and 0.538 for metaphor. The MTL ALBERT\textsubscript{large} model achieves an F1 of 0.312 (6.12\% $\uparrow$) for hyperbole and 0.475 (4.62\% $\uparrow$) for metaphor over the STL ALBERT\textsubscript{large} model with F1 of 0.294 for hyperbole and 0.454 for metaphor. The MTL RoBERTa\textsubscript{large} model achieves an F1 of 0.561 (13.10\% $\uparrow$) for hyperbole and 0.573 (16.93\% $\uparrow$) for metaphor over the STL RoBERTa\textsubscript{large} model with F1 of 0.496 for hyperbole and 0.490 for metaphor. 
Overall the RoBERTa\textsubscript{lg} MTL-F model achieves the best precision of $0.565$, recall of $0.587$, and F1 of $\textbf{0.573}$ ($\textbf{16.93}$\% $\uparrow$)  for metaphor detection and a p-value < $\textbf{0.0001}$.

\textbf{4. LCC results:}
The comparative analysis results for the label-balanced LCC dataset are in Table \ref{tab:lcc_result}. For all the models we observe that the MTL-F performs better than the corresponding STL for metaphor detection. 
% The MTL RoBERTa\textsubscript{large} model achieves an F1 of 0.663 (3.91\% $\uparrow$) for hyperbole and 0.803 (0.37\% $\uparrow$) for metaphor over the STL RoBERTa\textsubscript{large} model with F1 of 0.638 for hyperbole and 0.800 for metaphor. 
Overall the RoBERTa\textsubscript{lg} MTL-F model achieves the best recall of $0.812$, and F1 of $\textbf{0.805}$ ($1.38$\% $\uparrow$) for metaphor detection and a p-value of $\textbf{0.0221}$.

We observe: a) The MTL-F model helps in achieving generalization under the presence of both hyperbole and metaphor labels. b) The p-values (30 samples) suggest that the MTL-F results are statistically significant over the STL results with $95$\% confidence for all the datasets (\textbf{Appendix} \ref{hypothesis_testing}). 
% either metaphor or hyperbole for hyperbole or metaphor detection respectively.

\noindent\textbf{STL vs. MTL-E vs. MTL-F models} \hspace{1em}Table \ref{tab:final_result} reports the comparison of these three models on the HYPO and LCC datasets for hyperbole and metaphor detection respectively. We observe that, in comparison to the STL model, the MTL-E model performs better in general whereas the MTL-F model performs significantly better, achieving the best F1 score of $\textbf{0.881}$ and $\textbf{0.805}$ on the HYPO and LCC datasets respectively. (See \textbf{Appendix} \ref{appendix_comparison}).

% Detailed comparison of the STL, MTL-E and MTL-F models are reported in Table \ref{tab:extra_hypo_result} and Table \ref{tab:extra_met_result}. For hyperbole detection we used the HYPO dataset whereas for metaphor detection we used label balanced LCC dataset. 
% Table \ref{tab:final_result} in the paper reports the comparison of only the best performing models for brevity.

\noindent \textbf{Benchmark Results} \hspace{1em}We report the benchmark results for sentence-level detection on the label balanced LCC dataset in Table \ref{tab:final_result} (check the Metaphor column). Our RoBERTa\textsubscript{lg} MTL-F model achieves the best recall of $\textbf{0.812}$ and F1 of $\textbf{0.805}$.

\noindent\textbf{Baseline Comparison} \hspace{1em}Table \ref{Table:baseline comparison} reports the comparison of our work with baseline models on the HYPO dataset for hyperbole detection. Our RoBERTa\textsubscript{lg} MTL-F model achieves the best recall of $\textbf{0.884}$ ($8.59$\% $\uparrow$) and F1 of $\textbf{0.881}$ ($\textbf{12.03}$\% $\uparrow$) as compared to the recall of $0.814$ and F1 of $0.781$ of the state-of-the-art system. 
% We observe that metaphor labels help to improve the efficiency of hyperbole detection.

\section{Analysis}
% We divide our analysis in two parts: 1) Comparative analysis between the STL and MTL models. 2) Error analysis of the MTL model.
% We divide our analysis into two subsections: the first subsection compares the STL and MTL-F models, while the second section does an error analysis of the MTL-F model. 
We divide our analysis into two subsections: 1) A comparison of the STL and MTL-F models, and 2) Error analysis of the MTL-F model.
%\todo{To be added}
% \todo{Tejpal}
\subsection{Comparative Analysis}
Under similar experimental setups, we compare the STL and MTL-F models on example sentences obtained from the different test sets of the cross-validation run of the HYPO dataset as shown in Table \ref{tab:comparison_results}. We consider the following $4$ cases:

% Under similar experimental setups, we compare the STL and MTL models on example sentences as shown in Table \ref{tab:comparison_results}. We choose sentences across the different test sets obtained from the cross validation run of the HYPO dataset and analyze them.  We consider the following 4 cases:

% We choose the HYPO dataset for the comparative analysis as the experimental results for both hyperbole and metaphor detection were far better compared to other datasets. We consider the following 4 cases: We choose the HYPO dataset for the comparative analysis.

% \noindent
\textbf{1. Hyperbolic and Metaphoric:} 
% Here we take sentences which are both hyperbolic and metaphoric and compare the cases where STL fails in comparison to the MTL model. 
\textit{"They cooked a turkey the size of a cow,"} is both hyperbolic and metaphorical. Here, the exaggeration is evident as the size of the turkey is being compared to that of a cow, which allows both the STL and MTL-F models to make correct hyperbole predictions. 
% But, the MTL model due to the correct detection of hyperbole helps in catching the implicit meaning of the size being big which does not happen in the STL model.
However, for metaphor prediction, the MTL-F model correctly identifies the implicit meaning of "size being big" under the influence of the correct hyperbole label, while the STL model fails to do so.

Next, for the example sentence, \textit{"Your plan is too risky, it’s a suicide,"} the exaggeration and the metaphoricity are very intricate. The words \textit{risky} and \textit{suicide} make it difficult for the STL model to detect the labels, but the MTL-F model accurately identifies them. This can be attributed to the MTL-F model's ability to learn from both labels.

% \noindent
\textbf{2. Non-Hyperbolic and Non-Metaphoric:} In some cases, the STL model may incorrectly classify sentences that are non-hyperbolic and non-metaphoric due to ambiguous language.
% For the sentences which are both non-hyperbolic and non-metaphoric and are tagged incorrectly by the STL model, we observe the presence of words which might indicate ambiguity towards exaggeration or metaphoricity. 
For example, in the sentence \textit{"I’m not staying here any longer!"} the words \textit{staying} and \textit{longer} may give the impression of exaggeration, causing the STL model to incorrectly classify it as hyperbolic.  

However, the MTL-F model, by learning both hyperbole and metaphor detection simultaneously, is able to identify such cases as non-hyperbolic. Similarly, in \textit{"My ex boyfriend! Treacherous person!"} the word \textit{treacherous} may lead the STL model to incorrectly classify it as hyperbolic and metaphoric, but the MTL-F model classifies it correctly.

% \noindent
\textbf{3. Hyperbolic and Non-Metaphoric:} For this category, we notice that similes can cause confusion. For instance, in the sentence \textit{"This kind of anger rages like a sea in a storm,"} \textit{anger} is explicitly compared to \textit{sea in a storm} through the word \textit{like}. The MTL-F model is able to distinguish this as a simile, whereas the STL model fails to do so.

% In cases where the sentences are hyperbolic and non-metaphoric but the STL model tags them incorrectly we observe that there are some instances where the presence of simile might be creating the confusion. For example, in the sentence \textit{"This kind of anger rages like a sea in a storm."} anger is compared to the sea during a storm but the comparison is brought about through the word "like" which the MTL model learns to differentiate as a simile but the STL model fails to do so.

% \noindent
\textbf{4. Non-Hyperbolic and Metaphoric:} Here we observe that the use of figurative language is subtle. For instance, in \textit{"Her strength awoke in poets an abiding love,"} \textit{awoke} is used metaphorically, which is correctly identified by both the STL and MTL-F models. However, the STL model incorrectly tags it as hyperbolic, while the MTL-F model learns to identify such sentences as non-hyperbolic.

% we observe that there are some cases where the STL model mistakes the sentences as hyperbolic. Here the word "awoke" has been used in a metaphoric sense which is correctly detected by both the STL and MTL model but the STL model tags it as hyperbolic whereas the MTL model learns to identify such sentences as non-hyperbolic.

\noindent \textbf{Analysis of attention weights:}\\
Additionally, we also examine the attention weights from the final layer to gain an insight into the performance of the MTL-F model compared to the STL model. We use the weights associated with the \textbf{[CLS] / <s>} ([CLS] for BERT and <s> for RoBERTa) token normalized over all the attention heads. 

% The attention weight corresponding to these tokens are used during the classification and we normalize them over all the attention heads and then across all the tokens to obtain a probability distribution over them. 

First, we compare the STL and MTL-F models for the task of hyperbole detection. Figure \ref{fig:hyp_attention}. shows attention weight comparison of example sentences. For the sentence \textit{"Hope deferred makes the heart sick,"} we observe that the MTL-F model focuses on the words \textit{heart} and \textit{sick} that indicate exaggeration, while the STL model focuses on other irrelevant words. Similarly, for \textit{"Books are food for avid readers,"} the MTL-F model correctly focuses on the words \textit{Books}, \textit{food} and \textit{readers}. This suggests that the MTL-F model is better at paying attention to relevant words in the sentence due to its knowledge of both hyperbole and metaphor detection.

% the maximum attention of the STL model is on the tokens \textit{the} and \textit{heart} whereas for the MTL model it is on the tokens \textit{heart} and \textit{sick}. Here the MTL model correctly identifies the words that are source of exaggeration. 
% Similarly, for the sentence \textit{"Books are food for avid readers."} the STL model focuses on the tokens \textit{Books}, \textit{are}, and \textit{food} whereas the MTL model focuses on the tokens \textit{Books}, \textit{are}, \textit{food}, and \textit{readers}. We observe that the MTL models are able to attend to correct words in the sentences because they have insights into the hyperbole and metaphor classification.

% We observe that the MTL model thus due to better choice of tokens to attend to learns to tag such sentence as hyperbolic where the STL model fails to do so.

% Here it makes sense for the MTL model to pay attention to these words as they are the cause of exaggeration in this particular sentence. 

Next, for metaphor detection, the presence of hyperbole labels during training helps the MTL-F to learn to correctly attend to relevant tokens. For example, in \textit{"After workout I feel I could lift a sumo wrestler,"} the MTL-F focuses on the words \textit{lift} and \textit{wrestler} to correctly identify it as metaphoric. Similarly, for \textit{"Seeing my best friend again would mean the world to me,"} the MTL-F pays the maximum attention to the words \textit{would}, \textit{mean}, and \textit{world} which is the reason for metaphoricity here.

%is metaphoric as it compares the size of turkey to that of cow and is also exaggerated. The STL model correctly identifies the sentence as hyperbolic but incorrectly predicts the sentence as non-metaphoric whereas the MTL model is able to correctly identify the sentences as both hyperbolic and metaphoric. 

% \begin{figure*}
%     \centering
%     \begin{minipage}{0.3\textwidth}
%         % \hspace*{-1cm}%
%         \includegraphics[scale=0.3]{images/Hope deferred makes the heart sick.png}
%         % \hspace*{-1cm}%
%         \includegraphics[scale=0.3]{images/Books are food for avid readers.png}
%         \caption{Caption}
%         \label{fig:my_label}
%     \end{minipage}
%     \hfill
%     \begin{minipage}{0.7\textwidth}
%         % \hspace*{-1.5cm}%
%         \includegraphics[scale=0.3]{images/After workout I feel I could lift a sumo wrestler.png}
%         % \hspace*{-1.5cm}%
%         \includegraphics[scale=0.3]{images/Seeing my bestfriend again would mean the world to me.png}
%         \caption{Caption}
%         \label{fig:my_label}
%     \end{minipage}
    
% \end{figure*}

\subsection{Error Analysis}
We also analyzed the misclassifications for the MTL-F model, some of which have been included in Table \ref{tab:mtl_error}. We observe that the primary reason for misclassifications in the MTL-F model is the lack of context in identifying the exaggeration or metaphoricity. For instance, \textit{"What kind of sorcery is this?"} is a commonly used figurative sentence but the absence of any context makes it difficult for the MTL-F model to classify it correctly as both hyperbolic and metaphoric. 

Next, we found cases such as \textit{"You're grumpy,"} where the MTL-F model tags them incorrectly as metaphoric. Such mistakes could be attributed to the model learning to identify implicit comparisons but failing to identify that \textit{grumpy} here is an attribute not a comparison.
% like \textit{"He was a lion in the war,"} 
\section{Conclusion and Future work}
%\todo{need to change once other sections are done}
% In this paper, we proposed a multi-task deep learning framework to simultaneously detect both hyperboles and metaphors at sentence-level. Due to the absence of suitable datasets for this particular task, we manually annotated the hyperbole and metaphor datasets. We used different transformer models to perform our experiments and compared our results. Empirical evaluations show that the proposed multi-task approach achieves better performance on hyperbole and metaphor detection.

% We have presented a novel multi-tasking approach to the detection of hyperboles and metaphors. Our augmentation of annotation of two hyperbole datasets with metaphor labels and that of two metaphor datasets with hyperbole labels lead to multi-task learning of metaphor and hyperbole, which scores over single-task learning for both tasks.  We establish new SOTA for hyperbole detection and new benchmark for sentence-level metaphor detection. The take-away message is that metaphor and hyperbole detection help each other and should be done together.

We have presented a novel multi-tasking approach to the detection of hyperboles and metaphors. We augmented the annotations of two hyperbole datasets with metaphor labels and that of two metaphor datasets with hyperbole labels. This allowed multi-task learning of metaphor and hyperbole detection, which outperforms single-task learning on both tasks.  We establish a new SOTA for hyperbole detection and a new benchmark for sentence-level metaphor detection. The take-away message is that metaphor and hyperbole detection help each other and should be done together.

% We extend the existing hyperbole and metaphor datasets with manually annotated labels so that each sentence has both hyperbole and metaphor labels, across datasets.
% We show quantitative and qualitative results to support the hypothesis that metaphors aid in the detection of hyperbole and vice-versa.

% , which highlights the importance of studying these language phenomena together.

% Overall, this work contributes to the field of natural language processing by providing a new method for detecting hyperbole and metaphor in text and by highlighting the importance of studying their relationship.
We plan to extend our framework of exploiting linguistic relatedness and thereby creating MTL detection systems, to all forms of figurative languages like proverbs, idioms, humour, similes, and so on.

% In future, the multi-task framework can be expanded to include other figurative language like personification, idiom etc., and can also be used to convert such sentences to literal sentences which is essential in many NLP systems like chatbots and personal assistants.

% The knowledge gained to 

% The figurative language detection systems can be augmented to paraphrase the sentences to literal sentences.

% Future work includes expanding the multi-task framework for detection of other figurative language like personification, idiom etc. It also includes conversion of given hyperbole or metaphor sentence to a literal sentence that represents the same meaning. The proposed multi-task framework for hyperbole and metaphor detection is used to identify whether the given sentence is a hyperbole or metaphor or literal sentence. Once it is identified as hyperbole or metaphor, it is important to convert them to its corresponding literal sentence so that the current NLP systems understand the figurative language effectively. It has many use cases in dialogue systems where chatbots or personal assistants can process and respond to hyperbole or metaphor statements effectively.

\section{Limitations}
% In this work, we explored sentence-level detection of hyperbole and metaphor. However, the span-level detection of hyperbole or metaphor i.e. localizing to the span of words in the sentence which makes the sentence hyperbole or metaphor is important to understand them effectively. The limitation of proposed multi-task framework of hyperbole and metaphor detection assumes the other figure of speech other than hyperbole and metaphor as literal sentences.

%In this work, we explored sentence-level detection of hyperbole and metaphor due to the nature of datasets. We have not expanded our work to detect the words or phrases that convey the hyperbole and metaphor in the sentence as existing hyperbole datasets do not contain span level annotations. The proposed multi-task framework of hyperbole and metaphor detection assumes the sentences with other figurative languages like similes as literal sentences which will require a fine-grained annotation when dealing with a generalized figurative language detection framework.

The scope of this work is limited to sentence-level detection due to the absence of any span-level annotated datasets for hyperbole detection. Also, we could only partially annotate the metaphor datasets due to resource constraints. Finally, we did not try sophisticated large language models in our work as our goal was to demonstrate the effectiveness of multitasking using a simple model, rather than to test the performance of more sophisticated models.

\section{Ethical Considerations}
We perform our experiments on existing hyperbole and metaphor datasets by adding additional labels to them. Some of the examples in these datasets use slurs, abuses, and other derogatory terms to bring out exaggeration or implicit comparison. Our models may also propagate these unintended biases due to the nature of the datasets. We urge the research community to use our models and these datasets with caution and we are fully committed to removing discrepancies in the existing hyperbole and metaphor datasets in the future.

% We are fully committed to removing the discrepancies in these datasets and will consider making 

\section*{Acknowledgements}
We thank our annotation team for all the efforts taken to ensure quality during the annotation. Special mention to Akash Bhagat from the annotation team for his efforts. We thank the anonymous reviewers for their insightful feedback. We also thank our colleagues at CFILT Lab, IIT Bombay for their valuable comments and suggestions.

\bibliography{anthology,custom}
\bibliographystyle{acl_natbib}

\appendix

\section{Appendix}
\subsection{Experimental Details}\label{section:appendix experiment}
For experiments, we use the NVIDIA A100-SXM4-80GB GPU. Check Table \ref{tab:exp_add} for further details.
\begin{table}[h]
    \centering
    \begin{tabular}{|c|c|c|}
    \hline
       \textbf{Model} & \textbf{\#Parameters} &\textbf{Run time}\\
    \hline
    \hline
    % BERT\textsubscript{base} & 0.788 & 0.735 & 0.756 & 0.720 & 0.631 & 0.670\\
     \small{BERT\textsubscript{lg}} & $\sim335M$ & $\sim25$ mins \\
     \small{ALBERT\textsubscript{xxl2}} & $\sim222M$ & $\sim45$ mins 
     \\
     \small{RoBERTa\textsubscript{lg}} & $\sim355M$ & $\sim26$ mins  \\
    %  & deBERTa\textsubscript{large} & \textbf{0.899} & \textbf{0.887} & \textbf{0.892} & 0.797 & 0.741 & 0.766 \\
     \hline
     
    \end{tabular}
    \caption{Additional details of the models along with their number of parameters and run time.}
    \label{tab:exp_add}
\end{table}

\subsection{Label Balanced Metaphor Datasets}\label{label-balanced}
As discussed in Section \ref{section: experiments}, we used label-balanced metaphor datasets to address the imbalance caused by fewer hyperbole. Table \ref{tab:balanced_stats} reports the statistics of the label-balanced metaphor datasets. 
\begin{table}[h]
\centering
\begin{tabular}{|l|l|l|l|l|}
\hline
\textbf{Dataset} & \textbf{\#H} & \textbf{\#NH} & \textbf{\#M} & \textbf{\#NM} \\
\hline
\textbf{TroFi}   & 444          & 1100             & 709         & 835             \\
\hline
\textbf{LCC}     & 634          & 1400             & 1217        & 817  \\
\hline
\end{tabular}
\caption{Statistics of label balanced metaphor datasets. \#H, \#NH, \#M, and \#NM represent the number of hyperboles, non-hyperboles, metaphors, and non-metaphors respectively.}
    \label{tab:balanced_stats}
\end{table}

\subsection{STL vs. MTL-E vs. MTL-F models}\label{appendix_comparison}
% Comparisons of STL, MTL-E, and MTL-F models for hyperbole and metaphor datasets are reported
Detailed comparison of the STL, MTL-E and MTL-F models are reported in Table \ref{tab:extra_hypo_result} and Table \ref{tab:extra_met_result}. For hyperbole detection we used the HYPO dataset whereas for metaphor detection we used label balanced LCC dataset. Table \ref{tab:final_result} in the paper reports the comparison of only the best performing models for brevity.
\begin{table}[h]
    \centering
    \begin{tabular}{|c|l|c|c|c|}
    \hline
       \multirow{2}{*}{\textbf{Task}} & \multirow{2}{*}{\textbf{Model}} & \multicolumn{3}{c|}{\textbf{Hyperbole}} \\
    \cline{3-5}
        & &  \small{Precision} & \small{Recall} & \small{F1}\\
    \hline
    \hline
    \multirow{3}{*}{\small{STL}} 
    % BERT\textsubscript{base} & 0.788 & 0.735 & 0.756 & 0.720 & 0.631 & 0.670\\
     & \small{BERT\textsubscript{lg}} & 0.827 & 0.801 & 0.811\\
     & \small{ALBERT\textsubscript{xxl2}} & 0.845 & 0.871 & 0.856\\
     & \small{RoBERTa\textsubscript{lg}} & \textbf{0.883} & 0.848 & 0.864\\
    %  & deBERTa\textsubscript{large} & \textbf{0.899} & \textbf{0.887} & \textbf{0.892} & 0.797 & 0.741 & 0.766 \\
     \hline
      \multirow{3}{*}{\small{MTL-E}} 
    %  & BERT\textsubscript{base} & 0.818 & 0.785 & 0.792 & 0.790 & 0.645 & 0.706\\
      & \small{BERT\textsubscript{lg}} & 0.817 & 0.788 & 0.802\\
     & \small{ALBERT\textsubscript{xxl2}} & 0.726 & 0.746 & 0.734\\
     & \small{RoBERTa\textsubscript{lg}} & 0.859 & 0.878 & 0.867\\
     \hline
     
     \multirow{3}{*}{\small{MTL-F}} 
    %  & BERT\textsubscript{base} & 0.818 & 0.785 & 0.792 & 0.790 & 0.645 & 0.706\\
      & \small{BERT\textsubscript{lg}} & 0.853 & 0.824 & 0.836\\
     & \small{ALBERT\textsubscript{xxl2}} & 0.847 & 0.878 & 0.860\\
     & \small{RoBERTa\textsubscript{lg}} & 0.879 & \textbf{0.884} & \textbf{0.881}\\
    %  & deBERTa\textsubscript{large} & 0.892 & 0.886 & 0.887 & \textbf{0.825} & 0.724 & \textbf{0.769} \\
     \hline
     
    \end{tabular}
    \caption{Comparison of STL, MTL-E, and MTL-F models on the HYPO dataset for hyperbole detection.}
    \label{tab:extra_hypo_result}
\end{table}
\begin{table}[h]
    \centering
    \begin{tabular}{|c|l|c|c|c|}
    \hline
       \multirow{2}{*}{\textbf{Task}} & \multirow{2}{*}{\textbf{Model}} & \multicolumn{3}{c|}{\textbf{Metaphor}} \\
    \cline{3-5}
        & &  \small{Precision} & \small{Recall} & \small{F1}\\
    \hline
    \hline
    \multirow{3}{*}{\small{STL}} & \small{BERT\textsubscript{lg}} & 0.758 & 0.736 & 0.745\\
    & \small{ALBERT\textsubscript{xxl2}} & 0.723 & 0.757 & 0.739\\
    & \small{RoBERTa\textsubscript{lg}} & \textbf{0.802} & 0.787 & 0.794\\
    
     \hline
     \multirow{3}{*}{\small{MTL-E}} 
    %  & BERT\textsubscript{base} & 0.818 & 0.785 & 0.792 & 0.790 & 0.645 & 0.706\\
      & \small{BERT\textsubscript{lg}} & 0.737 & 0.762 & 0.749\\
     & \small{ALBERT\textsubscript{xxl2}} & 0.654 & 0.610 & 0.629\\
     & \small{RoBERTa\textsubscript{lg}} & 0.792 & 0.808 & 0.799\\
     \hline
     
     \multirow{3}{*}{\small{MTL-F}} & \small{BERT\textsubscript{lg}} & 0.750 & 0.774 & 0.760\\
    & \small{ALBERT\textsubscript{xxl2}} & 0.709 & 0.785 & 0.744\\
    & \small{RoBERTa\textsubscript{lg}} & 0.798 & \textbf{0.812} & \textbf{0.805}\\
     \hline
     
    \end{tabular}
    \caption{Comparison of STL, MTL-E, and MTL-F on the label balanced LCC dataset for metaphor detection.}
    \label{tab:extra_met_result}
\end{table}

\section{Case study: Hyperbole and Metaphor Detection using LLMs}

We tested the zero-shot capabilities of recent LLMs for hyperbole and metaphor detection. We experimented with a varied set of prompts on the open sources LLMs- LLaMA \cite{touvron2023llama} and BLOOM \cite{scao2022bloom}. We found that they do not have any deeper understanding of hyperboles and metaphors and performed poorly in the classification task.
Our observations with ChatGPT are as follows. ChatGPT had a good understanding of what hyperbole and metaphor are and was able to classify popular examples correctly. However, it struggled to perform multi-class classification when sentences had both hyperbole and metaphor properties. It also struggled to identify many hyperboles. When pointed out explicitly that it has made a mistake, and a particular part of the sentence is an indication of hyperbole or metaphor, it was able to correct itself and explain the reason. We experienced such issues with all combinations of presence or absence of metaphor and hyperbole.
The example instances demonstrating the same are shown below. \\
\subsection{Example 1}

\textit{\textbf{User}:	 Can you detect if the sentence has hyperbole or metaphor or both in it?
Sentence: That sharp bend is a death-trap for motorists.\\
\textbf{ChatGPT}:  Yes, I can detect that the sentence contains a metaphor. The sentence "That sharp bend is a death-trap for motorists" is a metaphor because it is comparing the sharp bend to a death-trap, suggesting that the bend is extremely dangerous and potentially deadly for motorists. The metaphor is used to create a vivid and impactful image in the reader's mind.\\
\textbf{User}: Does it have a hyperbole? \\
\textbf{ChatGPT}:  No, the sentence does not contain hyperbole. Hyperbole is an exaggerated statement or claim not meant to be taken literally. The sentence "That sharp bend is a death-trap for motorists" is not an exaggeration, but rather a metaphorical comparison between the sharp bend and a death-trap.\\
\textbf{User}: Isn't "death-trap" a hyperbole?\\
\textbf{ChatGPT}: You are correct, "death-trap" in this context can be considered a hyperbole…..}

\subsection{Example 2}

\textit{\textbf{User}:	Can you detect if the sentence has hyperbole or metaphor in it? Sentence: Retailers, meanwhile, are trying out a number of ways to target these shoppers.\\
\textbf{ChatGPT}: The sentence "Retailers, meanwhile, are trying out a number of ways to target these shoppers" does not contain hyperbole or metaphor. It is a straightforward statement conveying information about the actions of retailers in targeting a particular group of shoppers. \\
\textbf{User}: Isn't the word "target" a metaphor here?\\
\textbf{ChatGPT}: You are correct that "target" in this sentence can be considered a metaphor. In this context, "target" is being used figuratively to mean that retailers are trying to aim their marketing efforts towards a particular group of shoppers. The word "target" is being used in a non-literal sense to describe the retailers' actions. Therefore, the sentence does contain a metaphor. Thank you for pointing that out.}

It can be seen that in both examples, the model initially makes the wrong assumption about the sentence being a hyperbole or metaphor. It was able to correct itself only after bringing attention to the important word in the sentence. We have shown that the correct words get more attention through our multi-tasked approach indicating the reason for better detection accuracy.

\end{document}